\documentclass[runningheads]{ola}

\usepackage{epsfig}
\usepackage{amsmath}
\usepackage{amssymb} 
\usepackage{graphics}
\usepackage{psfrag}
\usepackage{epstopdf}
\usepackage{balance}
\usepackage{array}
\usepackage{caption}
\usepackage{multirow}
\usepackage{tabularx}
\usepackage{diagbox}
\usepackage[linesnumbered,ruled]{algorithm2e}
\renewcommand{\figurename}{Fig.}

\usepackage{calligra}
\DeclareMathAlphabet{\mathcalligra}{T1}{calligra}{m}{n}

\begin{document}

\pagestyle{headings}

\mainmatter

\title{Statistical learning for sensor localization\\ in wireless networks}

\titlerunning{Statistical learning for sensor localization in wireless networks}

\author{Daniel Alshamaa\inst{1}, Farah Mourad-Chehade\inst{1}, \and Paul Honeine\inst{2}}

\institute{Institut Charles Delaunay, ROSAS, LM2S, Universit\'e de Technologie de Troyes, Troyes, France \\
\email{daniel.alshamaa@utt.fr, farah.chehade@utt.fr}\\
\and
LITIS lab, Universit\'e de Rouen, Rouen, France \\
\email{paul.honeine@univ-rouen.fr}
}

\maketitle

\section{Introduction}

Indoor localization has become an important issue for wireless sensor networks \cite{ieee_sensors_journal}, \cite{localization_in_malls}, \cite{loc_in_museums}. This paper presents a zoning-based localization technique that uses WiFi signals and works efficiently in indoor environments. The targeted area is composed of several zones, the objective being to determine the zone of the sensor using an observation model based on statistical learning.

\section{Localization approach}

\subsection{Problem description}

The localization problem is tackled in the following manner. Let $N_{\!Z}$ be the number of zones of the targeted area, denoted by $Z_k,~ k=1,2,\ldots,N_{\!Z}$ and $N_{\!A\!P}$ be the number of detected Access Points (APs), denoted by $A\!P_n,~ n=1,2\ldots,N_{\!A\!P}$. The aim of the presented algorithm is to propose an observation model, using the WiFi received signal strength (RSS) \cite{eusipco}, to assign a confidence level $\mathcal{C}\!f(Z_k)$ for each zone $Z_k$, for any new observation $\boldsymbol{\rho}$. Here, $\boldsymbol{\rho}$ is a vector of size $N_{\!A\!P}$ of RSS measurements collected by the sensor from surrounding APs. The observation model uses RSS data received from surrounding WiFi APs to estimate the zone of the sensor. In an offline phase, fingerprints are collected by measuring the RSS of all existing APs in random positions of each zone. Then, in the online phase, once a new measurement of RSS is received, the model is used to assign a certain confidence to each zone in a belief functions framework.

      
\begin{figure}[ht]
\begin{minipage}[b]{0.45\linewidth}
\subsection{Statistical learning}

The RSS measurements of each zone are fitted to distributions defined over a set of parameters to be estimated with the available data. First, we choose the  types of distributions to be tested. Then, we estimate their parameters using the observations. And finally, we apply a statistical goodness of fit test to evaluate their fitting error. The problem is in the form of hypothesis testing where the null and alternative hypotheses are:

$H_0$: Sample data come from the stated distribution.

$H_a$: Sample data do not come from the stated distribution.
\end{minipage}
\hspace{0.5cm}
\begin{minipage}[b]{0.45\linewidth}
\centering
\includegraphics[width=1.1\textwidth]{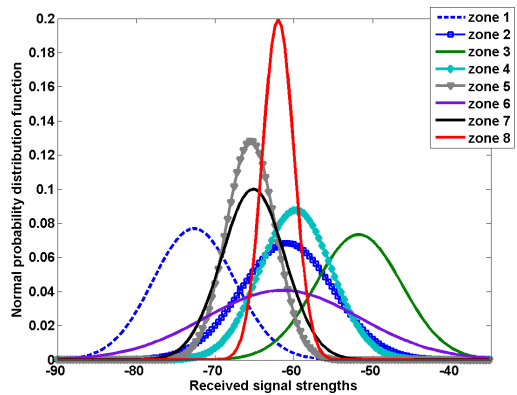}
\caption{Fitted distributions representing eight zones with respect to an AP.}
\label{fig:normal}
\end{minipage}
\end{figure}
\noindent The Kolmogorov-Smirnov (K-S) test \cite{method} is used to test the hypotheses. For each considered distribution, the hypothesis $H_0$ is rejected at a significance level $\alpha$ if the test statistic is greater than a critical value  obtained from the K-S table \cite{kolmogorov_smirnov_table}. All the considered distributions are tested, and the accepted ones are ranked according to their statistics, the best fitting one being selected \cite{spawc18}. Figure \ref{fig:normal} shows an example of fitting Gaussian distributions to eight zones with respect to an AP. 

\subsection{Online localization}

Let $\mathcal{Z} $ $=\{Z_1,\ldots,Z_{N_{\!Z}}\}$ be the set of all possible zones and let $2^\mathcal{Z}$ be the set of all the supersets of $\mathcal{Z}$, i.e., $ 2^\mathcal{Z}= \{\emptyset, \{Z_1\}, \ldots, \mathcal{Z}\}$. The cardinal of $2^\mathcal{Z}$ is equal to $2^{|\mathcal{Z}|}=2^{N_{\!Z}}$, where $|\mathcal{Z}|$ denotes the cardinal of $\mathcal{Z}$. One fundamental function of the BFT is the mass function, also called the basic belief assignment (BBA). The mass $m_{A\!P_n}(A)$ given to $A\in 2^{\mathcal{Z}}$ stands for the proportion of evidence, brought by the source $A\!P_n$, saying that the observed variable belongs to $A$.\\
In order to define the APs BBAs, all observations related to each AP belonging to a superset $A\in 2^{\mathcal{Z}}$ are fitted to a distribution $Q_{A\!P_n,A}$ \cite{ntms_localization}. Then, having an observation $\rho_{n}$ related to $A\!P_n, n\in \{1,\ldots,N_{\!A\!P}\}$, the mass $m_{A\!P_n}(A)$ is calculated as follows,
\begin{equation}\label{eq:mass}
m_{A\!P_n}(A)=\frac{Q_{A\!P_n,A}(\rho_{n})}{\sum_{A'\in 2^{\mathcal{Z}}, A'\neq \emptyset} Q_{A\!P_n,A'}(\rho_{n})},~~A \in 2^{\mathcal{Z}}, A\neq \emptyset.
\end{equation}
The quantity $m_{A\!P_n}(A)$ represents the amount of evidence brought by the source $A\!P_n$ saying that the observation $\rho_{n}$ belongs to the set $A$, $A$ being a singleton, a pair, or more. By taking all the supersets of ${\mathcal{Z}}$ and not only the singletons, the proposed algorithm uses all available evidences, even if they are uncertain about a single element. 

\subsubsection{Fusion of evidence}
According to the information retrieved from the APs, the mass functions $ m_{A\!P_n}(\cdot)$ are defined. Combining the evidence consists in aggregating the information coming from all the APs \cite{daniel}. The mass functions can then be combined using the conjunctive rule of combination as follows,
\begin{equation}\label{eq:m1_2}
m_{\cap}(A) =\!\!\!\!\!\!\!\!\!\! \sum _{\substack{A^{(n)}\in 2^{\mathcal{Z}} \\ \cap_{n}A^{(n)} = A}}\!\!\!\!\!\!\!\!\!\!\!\!~~ m_{A\!P_1}(A^{(1)})\times ... \times  m_{A\!P_{N_{\!A\!P}}}(A^{(N_{\!A\!P})}), 
\end{equation}
for all the sets $A\in 2^{\mathcal{Z}}$, with $A^{(n)}$ is the set $A$ with respect to the Access Point $A\!P_n$. The mass function is then normalized, leading to the Dempster rule of combination:
\begin{equation}\label{eq:normalize}
m_{\bigoplus}(A)=\frac{m_{\cap}(A)}{\sum_{A'\in 2^{\mathcal{Z}}}m_{\cap}(A')}
\end{equation}

\subsubsection{Pignistic transformation}
An adequate notion of the BFT to attribute masses to singleton sets is the pignistic level \cite{decentralized}. It is defined as follows,
\begin{equation}\label{eq:pignistic}
BetP(A) =  \sum_{A\subseteq A'} \frac{m_{\bigoplus}(A')}{|A'|},
\end{equation} 
where $A$ is a singleton of $2^{\mathcal{Z}}$. The pignistic level is equivalent to the probability of having the observation belonging to the considered set. One could also compute the pignistic level of higher-cardinal supersets. However, only  the singleton sets are taken into consideration, as we are interested in determining a level of confidence for the original zones only. Hence, the confidence associated to each zone by the observation model can be computed as follows \cite{sensors},
\begin{equation}\label{eq:mass_assoc}
\mathcal{C}\!f(Z_k)=BetP(\{Z_k\}), k\in \{1,\ldots,N_{\!Z}\}.
\end{equation} 
The zone having the highest confidence is chosen as the zone where the sensor resides. 

%
%
%
%
%
\bibliographystyle{ieeetr}
\bibliography{ola}

\end{document}